\pdfoutput=1

\documentclass[11pt]{article}

\usepackage[final]{acl}

\usepackage{times}
\usepackage{latexsym}

\usepackage[T1]{fontenc}

\usepackage[utf8]{inputenc}

\usepackage{microtype}

\usepackage{inconsolata}

\usepackage{graphicx}

\usepackage{amsmath} 
\usepackage{amsfonts} 
\usepackage{graphicx}
\usepackage{multirow}
\usepackage{booktabs}
\usepackage{makecell}
\usepackage{cleveref}

\usepackage[utf8]{inputenc}
\usepackage{listings}
\usepackage{xcolor}
\usepackage{caption}
\usepackage{tcolorbox}
\tcbuselibrary{listingsutf8}
\usepackage{multicol}
\usepackage{geometry}
\usepackage{array}
\usepackage{booktabs}

\definecolor{codegreen}{rgb}{0,0.6,0}
\definecolor{codeblue}{rgb}{0.1,0.1,0.8}
\definecolor{backgray}{rgb}{0.95,0.95,0.95}
\lstdefinestyle{code}{
    backgroundcolor=\color{white},
    commentstyle=\color{codegreen},
    keywordstyle=\color{codeblue},
    numberstyle=\tiny\color{gray},
    stringstyle=\color{magenta},
    basicstyle=\ttfamily\scriptsize,
    breaklines=true,
    frame=none,
    showstringspaces=false,
    tabsize=1,
    language=Python
}
\lstdefinestyle{plain}{
    backgroundcolor=\color{white},
    basicstyle=\ttfamily\scriptsize,
    breaklines=true,
    frame=none,
    showstringspaces=false,
    tabsize=1,
    language=,            
    keywordstyle=,        
    stringstyle=,         
    commentstyle=,        
    numberstyle=,         
}

\newcommand{\stitle}[1]{\vspace{1ex} \noindent{\bf #1}}

\title{Code Execution as Grounded Supervision for LLM Reasoning}

\author{
Dongwon Jung\textsuperscript{1} \quad
Wenxuan Zhou \textsuperscript{2} \quad 
Muhao Chen\textsuperscript{1} \\
\textsuperscript{1}University of California, Davis,
\textsuperscript{2}University of Southern California, \\
\texttt{\{dwojung,muhchen\}@ucdavis.edu} \quad
\texttt{zhouwenx@usc.edu} 
}

\begin{document}

\maketitle

\begin{abstract}
Training large language models (LLMs) with chain-of-thought (CoT) supervision has proven effective for enhancing their reasoning abilities. However, obtaining reliable and accurate reasoning supervision remains a significant challenge. We propose a scalable method for generating a high-quality CoT supervision dataset by leveraging the determinism of program execution. Unlike existing reasoning dataset generation methods that rely on costly human annotations or error-prone LLM-generated CoT, our approach extracts verifiable, step-by-step reasoning traces from code execution and transforms them into a natural language CoT reasoning. Experiments on reasoning benchmarks across various domains show that our method effectively equips LLMs with transferable reasoning abilities across diverse tasks. Furthermore, the ablation studies validate that our method produces highly accurate reasoning data and reduces overall token length during inference by reducing meaningless repetition and overthinking.\footnote{Code and data are available at \url{https://github.com/luka-group/Execution-Grounded-Reasoning}}

\end{abstract}

\begin{figure*}[t]
    \centering
\small\includegraphics[width=\linewidth]{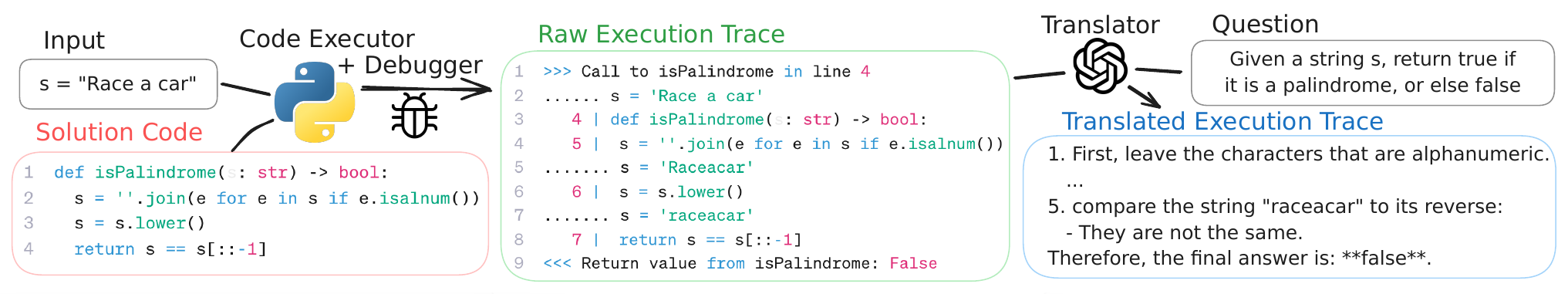}
    \caption{An overview of our method. The translated execution trace is grounded in code execution, making it a reliable and accurate source of reasoning supervision for the LLM.}
    \label{fig:method}
\vspace{-1em}
\end{figure*}

\section{Introduction}

Large language models (LLMs) have demonstrated strong performance across a range of complex reasoning tasks.
A key development in this area is chain-of-thought (CoT) training, which enhances LLMs by encouraging the generation of intermediate reasoning steps before producing a final answer \cite{JMLR:v25:23-0870,Ho2022LargeLM,Magister2022TeachingSL,Li2023SymbolicCD}.
CoT supervision has proven especially effective in improving the generalization and interpretability of LLMs, and has become a central component in the development of reasoning models \citep{ye2025limo,muennighoff2025s1,sky_t1_2025,Chang2025DemystifyingLC}.

Despite its success, obtaining high-quality CoT data at scale remains a major challenge for supervising the reasoning LLMs.
Existing CoT datasets are typically constructed in two ways. First, human-annotated CoT examples \citep{JMLR:v25:23-0870,cobbe2021training} provide high-quality and accurate reasoning guidance but are costly to acquire and non-scalable.
Second, many recent efforts rely on bootstrapped CoT data generated by prompting existing LLMs \citep{Magister2022TeachingSL,muennighoff2025s1}.
However, these synthetic data often suffer from intermediate reasoning errors, inconsistencies, and lack of grounding~\cite{zheng2024processbench,lyu2023faithful,chen2025reasoning}.
Although these methods verify and filter the CoT data at either the process or outcome level \citep{zelikman2022star,lightman2023let,luo2024improve,Li2025CodeIOCR}, they still fall short in guaranteeing the correctness of intermediate reasoning steps, undermining the reliability of the supervision signal.

In this work, we propose a scalable method for generating verifiable CoT data to supervise the reasoning process of LLM by leveraging the determinism of program execution.
Our core insight is that when problems can be formalized and solved with executable code, the resulting execution traces \cite{aakerblom2014tracing} provide inherently correct, step-by-step reasoning aligned with the task.
These traces offer a verifiable and error-free alternative to LLM-generated CoTs and can serve as a trustworthy source of supervision.

Specifically, we begin by sourcing open-source Python programs and executing them with a debugger to extract rich execution traces, including intermediate variable values, line-level execution order, and program control flow. Since the resulting raw execution traces lack natural language reasoning structures, we employ LLMs to translate the raw execution traces into fluent, human-readable rationales that resemble natural CoT data, effectively combining the correctness guarantees of execution with the expressive power of LLMs. 
Our method offers a scalable, annotation-free pipeline for generating high-quality and accurate reasoning supervision.

Experiments show that LLMs trained with our method demonstrate its effectiveness, achieving robust performance across coding, math, and reasoning tasks compared to baseline approaches. Ablation studies further confirm that our method improves data quality and reduces overall token length by mitigating meaningless repetition and overthinking.


\section{Method}

\subsection{Problem Settings}

Our goal is to enhance the reasoning capabilities of LLMs by supervising them with accurate and verifiable CoT reasoning traces.
Formally, given an input sequence $\mathbf{x} = [x_1, \ldots, x_m]$, an LLM $p_{\boldsymbol{\theta}}$ generates an output sequence $\mathbf{y} = [y_1, \ldots, y_n]$ through a sequence of intermediate reasoning steps $\mathbf{s} = [s_1, \ldots, s_l]$.
The overall generation process is defined as:
\[
p_{\boldsymbol{\theta}}(\mathbf{y}|\mathbf{x}) = p_{\boldsymbol{\theta}}(\mathbf{y}|\mathbf{s}, \mathbf{x}) \prod_{t=1}^{l} p_{\boldsymbol{\theta}}(s_t|\mathbf{s}_{<t}, \mathbf{x}), 
\]
where the model first generates each reasoning step $s$ conditioned on the input $\mathbf{x}$ and previous steps $\mathbf{s}_{<t}$, followed by generating the final answer $\mathbf{y}$ based on the full reasoning trace $\mathbf{s}$ and the original input $\mathbf{x}$.


High-quality CoT data is crucial for enabling strong reasoning performance in LLMs~\cite{lightman2023let}.
To collect CoT data at scale, existing approaches adopt a \emph{generate-then-filter} paradigm: they first sample CoTs using LLMs and then filter out low-quality ones.
Outcome-level filtering typically checks whether the final answer $\mathbf{y}$ matches the ground truth~\cite{xiong2025minimalist}, but this can miss flawed reasoning that coincidentally produces correct answers. 
Process-level filtering is more desirable as it evaluates intermediate reasoning quality, but remains challenging for current LLMs~\cite{zheng2024processbench}.


In this work, we propose a fundamentally different approach: constructing CoT data from code execution traces, which are inherently step-by-step, accurate, and causally linked to the final outcome, making them a natural source of accurate and verifiable reasoning supervision.
In the following sections, we describe how we construct high-quality CoT data, which is then used to fine-tune the LLM via supervised fine-tuning.

\subsection{Execution Trace Acquisition}

To efficiently obtain reliable and accurate CoTs, we leverage the abundance of coding data, which provide supervision in the form of problem–solution code pairs.
These pairs allow us to ground reasoning supervision in executable programs that reflect correct problem-solving logic.
Specifically, given a solution code snippet $\mathbf{c}$ and an input $\mathbf{x} = [\mathbf{x}_q; \mathbf{x}_i]$, where $\mathbf{x}_q$ is a natural language problem description and $\mathbf{x}_i$ is a concrete test input (e.g., a specific string or numerical input), we execute $\mathbf{c}$ using an execution tracing tool to obtain the returned answer $\mathbf{y}$ and a detailed execution trace $\mathbf{s}_{\text{trace}}$:
\begin{equation*}
\mathbf{y}, \mathbf{s}_{\text{trace}} = \texttt{Code\_Executor}(\mathbf{c}, \mathbf{x}_i). 
\end{equation*}
We employ a Python debugging tool called \texttt{Snoop} \citep{snoop} as the execution tracing tool, which records detailed line-by-line execution signals, including function calls and returns, executed lines of code, and the updated local variable values. An example of a \texttt{Snoop}-generated execution trace is provided in \Cref{fig:method}.\footnote{A more illustrative example of an execution trace is provided in \Cref{sec:execution_trace_example}} 
The resulting dataset is denoted as $D_{\text{trace}} = (\mathbf{x}, \mathbf{s}_{\text{trace}}, \mathbf{y})$,  
which contains verifiably accurate execution trace $\mathbf{s}_{\text{trace}}$ and the correct final output $\mathbf{y}$ for each instance, grounded in a verifiable code executor.

\subsection{Execution Trace Translation}
Although the execution trace $\mathbf{s}_{\text{trace}}$ captures the verifiable, step-by-step problem solving logic, its format differs substantially from natural language CoT reasoning.
Therefore, directly fine-tuning on such traces may hinder generalization and risks catastrophic forgetting on other reasoning tasks.
To better align the supervision signals with natural language reasoning, we transform the raw trace $\mathbf{s}_{\text{trace}}$ into a natural language CoT  $\mathbf{s}_{\text{nl\_trace}} = (s_1, ..., s_m)$ using an LLM as a \texttt{Translator}:
\begin{equation*}
\mathbf{s}_{\text{nl\_trace}} = \texttt{Translator}(\mathbf{x}, \mathbf{s}_{\text{trace}}).
\end{equation*}
The translator is prompted to emulate a human solving the problem by mentally tracing the code. It is instructed to express each reasoning step in natural language while faithfully reflecting the exact values and logic observed during code execution. This ensures that the output mirrors the precise reasoning behind the program’s behavior—grounded in execution, yet phrased as natural, intuitive, step-by-step thinking.
The resulting dataset, $D_{\text{nl\_trace}} = {(\mathbf{x}, \mathbf{s}_{\text{nl\_trace}}, \mathbf{y})}$, provides high-quality CoT traces that are both verifiable and linguistically aligned with typical LLM data, making them ideal for supervised fine-tuning.

\section{Experiment}
To assess whether supervision from code execution traces genuinely enhances reasoning ability, we conduct comprehensive experiments by comparing our approach with baseline datasets specifically designed to improve the reasoning capabilities of LLMs. In this section, we present the experimentation details and discuss the results.

\begin{table*}[t]
\centering
\scalebox{0.9}{
\begin{tabular}{lcccccccc}
\toprule
\multirow{2}{*}{\textbf{Methods}} & \multicolumn{3}{c}{\textbf{LiveBench}} & \multirow{2}{*}{\textbf{MATH500}} & \multirow{2}{*}{\textbf{BBH}} & \multirow{2}{*}{\textbf{AGIEval}} & \multirow{2}{*}{\textbf{GPQA}} & \multirow{2}{*}{\textbf{Avg}} \\
\cmidrule(r){2-4}
& \textbf{Code} & \textbf{Math} & \textbf{Reasoning} & \\
\midrule
No Training & 43.2 & 59.5 & 64.2 & 86.4 & 75.5 & 31.6 & 36.3 & 56.7 \\
Code & 28.5 & 33.1 & 51.7 & 82.4 & 69.7 & 31.2 & 33.9 & 47.2 \\
Raw Trace & 8.3 & 3.1 & 38.0 & 84.8 & 63.5 & 30.2 & 27.0 & 36.4 \\
CodeI/O & 39.3 & \textbf{62.7} & 62.2 & \textbf{86.6} & 81.3 & \textbf{32.4} & 33.9 & 56.9 \\
Ours & \textbf{44.5} & 61.0 & \textbf{65.8} & 86.4 & \textbf{81.4} & \textbf{32.4} & \textbf{38.1} & \textbf{58.5} \\
\bottomrule
\end{tabular}}
\caption{Experiment results (accuracy) of $\texttt{Qwen3-4B}$. Bolded scores indicate the highest performance.}
\label{tab:main}
\vspace{-1em}
\end{table*}

\subsection{Experiment Setup}

\paragraph{Data Generation} 

We select PyEdu-R, a subset of data from the Python-Edu \citep{benallal2024smollmcorpus} as the source of supervision dataset.
PyEdu-R focuses on STEM-related problems such as logic puzzles, math-related tasks, scientific computation, and system modeling. 
Since the original data only contains code, we utilize the preprocessed version that contains LLM-generated problem and the input-ouput pairs, made publicly available by \citet{Li2025CodeIOCR}. We obtain execution traces by running the code on the inputs and then translate these traces using $\texttt{Qwen3-32B}$ as the $\texttt{Translator}$. This process yields approximately 15K data instances.\footnote{We also include more details on data generation in \Cref{sec:code_exec_filter} and \Cref{sec:data_gen_config}}





\paragraph{Baselines}

We compare our method against several baselines designed to enhance LLM reasoning through fine-tuning. The training setup remains the same, with the only difference being the dataset curation process from the source data.
The compared baselines are:
(1) \textbf{No Training}, where the base LLM is evaluated without further fine-tuning;
(2) \textbf{Code Generation}, where the model is trained to generate solution code $\mathbf{c}$ given a question $\mathbf{x}_{\text{q}}$;
(3) \textbf{Raw Execution Trace}, where the model learns to generate the raw execution trace $\mathbf{s}_{\text{trace}}$ directly from the code $\mathbf{c}$ and input $\mathbf{x}_{\text{i}}$, bypassing the natural language translation; and (4) \textbf{CodeI/O}~\citep{Li2025CodeIOCR}, where the model is trained on CoT traces $\mathbf{s}_{\text{teacher}}$ produced by a teacher model, followed by a binary output correctness feedback.
For consistency, we use the same model employed in our \texttt{Translator} as the teacher in this baseline.

\paragraph{Models}

We conduct SFT on our data using two target models: \texttt{Qwen3-4B} and \texttt{Qwen3-8B}. We use $\texttt{Qwen3-32B}$ as both the translator in our method and the teacher model for the CodeI/O baseline. For both methods, we enable the $\texttt{enable\_thinking=True}$ option and extract the output after the thinking phase for the translation and the CoT generation results.

\paragraph{Evaluation Benchmark}

We evaluate our method and the baselines on widely adopted reasoning benchmarks including MATH500 \citep{lewkowycz2022solving}, BBH \citep{suzgun2022challenging}, AGIEval \citep{zhong2023agieval}, and GPQA \citep{rein2024gpqa}.
Additionally, we utilize LiveBench \citep{whitelivebench}, a recent comprehensive benchmark that contains diverse categories of tasks. We focus on the math, coding, and reasoning categories, as our primary goal is to evaluate the reasoning capabilities of the methods. Specifically, we use the 2024-11-25 release, which is currently the most up-to-date version.

\subsection{Experiment Results}
\label{sec:experiment_results}
The experiment results of $\texttt{Qwen3-4B}$ is presented in \Cref{tab:main}.\footnote{The experiment result of $\texttt{Qwen3-8B}$ is presented in \Cref{tab:main_appendix}} Overall, our method outperforms all baselines, demonstrating its effectiveness. 
The Code Generation and Raw Execution Trace baselines perform poorly across the board. Although both are derived from code-based supervision, they suffer from a misalignment with natural language reasoning formats. In particular, training on raw traces or code generation alone does not equip the model with the ability to produce step-by-step reasoning in natural language, leading to limited transfer and degraded performance—especially in reasoning-heavy tasks.

CodeI/O baseline remains strong on mathematical reasoning but suffers from a significant drop in performance on coding and science domains. This is likely because the reasoning structure of the teacher model, which is primarily optimized for math, is effectively distilled into the student model—potentially at the cost of performance in other domains. In contrast, our method achieves balanced improvements across all reasoning domains.

\subsection{Ablation Studies}
In this section, we present ablation studies evaluating the quality of our data and its effectiveness in reducing repetition and overthinking during inference.

\stitle{Better Quality Data.}
To assess the quality of the data generated by our method, we evaluate the correctness of (1) final output and (2) intermediate reasoning steps. We provide $\texttt{Qwen3-32B}$ with the generated solution and the ground truth answer to determine if it reaches the correct answer. For evaluating intermediate reasoning steps, we randomly select 200 samples and use a strong reasoning model, $\texttt{OpenAI-o3}$, to identify any errors within the reasoning process. 

\Cref{tab:output_correctness} shows the accuracy of both the final outputs and intermediate reasoning steps for our method and CodeI/O. Our method achieves higher accuracy than CodeI/O on both metrics, with a particularly larger margin in intermediate step accuracy. This demonstrates that our method produces more accurate reasoning steps, leading to the correct final output, as it is grounded in reliable code execution.

\stitle{Reducing Repetition and Overthinking.}
To evaluate the generation efficiency of the models trained on our method, we compare token lengths in \Cref{tab:token_length}. We show that models trained with our approach generate approximately 20\% fewer tokens for \texttt{Qwen3-4B} and 30\% fewer tokens for \texttt{Qwen3-8B}, compared to No Training baseline.\footnote{We include \texttt{Qwen3-8B} results in \Cref{tab:token_length_appendix}} Moreover, our method substantially reduces instances where the model reaches the maximum token limit due to overthinking or repetitive output. Importantly, this reduction in token length does not compromise performance, as demonstrated in \Cref{sec:experiment_results}.

\subsection{Case Study}
To further examine our data, we present two examples of reasoning trace in \Cref{tab:case_study1} and \Cref{tab:case_study2}. The first example demonstrates a case where the CodeI/O solution produces the correct final output despite an error in an intermediate step—specifically, it incorrectly lists the permutation of "hrf" in step 2—while our solution correctly completes all intermediate steps. The second example shows a case where both the intermediate reasoning and the final output are incorrect in the CodeI/O solution, due to an incorrect formula used at the beginning. In contrast, our solution produces correct calculations throughout, guided by the execution trace.

\begin{table}[t]
\centering
\scalebox{0.9}{
\begin{tabular}{lcc}
\toprule
\textbf{Method} & \textbf{Output} & \textbf{Intermediate} \\
\midrule
CodeI/O & 87.3 & 73.0 \\
Ours    & 98.3 & 91.5 \\
\bottomrule
\end{tabular}}
\caption{Comparison of correctness accuracy of final output and intermediate reasoning steps on LiveBench.}
\label{tab:output_correctness}
\end{table}
\begin{table}[t]
\centering
\scalebox{0.9}{
\begin{tabular}{lcc}
\toprule
\textbf{Method} & \textbf{Avg Token} & \textbf{Max Token Reached} \\
\midrule
No Training & 8804 & 123 \\
CodeI/O & 7684 & 91 \\
Ours    & 7068 & 72 \\
\bottomrule
\end{tabular}}
\caption{Token length statistics on the LiveBench evaluation using \texttt{Qwen3-4B}.}
\label{tab:token_length}
\vspace{-1em}
\end{table}

\section{Related Works}
\stitle{Reasoning Distillation from Teacher Models}
A common approach to distill reasoning ability is supervised fine-tuning on reasoning chains from stronger teacher models \citep{Ho2022LargeLM,Magister2022TeachingSL,Li2023SymbolicCD}. 
More recent work leverages test-time scaling to transform instruction-tuned models into Meta Chain-of-Thought (Meta-CoT;  \citealt{Xiang2025TowardsS2}), which first generate thought tokens before solving the problem. 
Our work is orthogonal to these approaches that leverage test-time scaling, and aims to improve the model’s inherent step-by-step reasoning ability.

\stitle{Training on Code for Reasoning}
Early studies have shown that LLMs trained on code excel at various reasoning tasks, including commonsense reasoning \citep{madaan2022language}, causal reasoning \citep{liu2023magic,liu2024llms}, and mathematical reasoning \citep{azerbayev2023llemma,shao2024deepseekmath}. However, these findings are limited to models heavily pre-trained on code and do not deeply investigate how code semantics and structure influence reasoning abilities. 

Recent studies have leveraged code execution traces to enhance reasoning on coding tasks, focusing on applications like vulnerability detection, program repair, and code generation \citep{ding2024traced,ni2024next,ding2024semcoder}. In contrast, our work targets task-agnostic, step-by-step reasoning that extends beyond the code domain.

The most closely related work is by \citet{Li2025CodeIOCR}, who train models on input–output prediction tasks using CoT rationales. While both approaches leverage CoT for output prediction, our method uses execution traces as grounded supervision, whereas theirs relies solely on binary output correctness. 

\section{Conclusion}
We introduce a 
novel approach that leverages code execution traces as verifiable and easily scalable supervision for enhancing step-by-step reasoning in LLMs. Our method incorporates both the ensured correctness of execution with the natural CoT reasoning steps to provide LLMs with high-quality reasoning supervision. Experiments across reasoning tasks show that our approach outperforms prior distillation methods, offering a reliable path toward improving LLM reasoning with grounded, annotation-free supervision.

\section*{Acknowledgment}

We appreciate the reviewers for their insightful
comments and suggestions.
This work was partly supported by the Amazon Nova Trusted AI Prize, the NSF of the United States Grants ITE 2333736 and OAC 2531126, and the DARPA FoundSci Grant HR00112490370.

\section*{Limitations}
While our method provides grounded and reliable reasoning supervision, it is inherently limited to tasks that can be expressed and solved via executable code. However, we have proved that our reasoning data transfers well to other domains like math and logical reasoning. Additionally, since the translation relies on an LLM, it may not always be perfect. Nonetheless, we have demonstrated the quality of our data by evaluating both the final outputs and the intermediate reasoning steps.

\section*{Ethics Statement}
This work follows the ACL Code of Ethics. We believe no potential risk is directly associated with the presented work.

\bibliography{reference}

\appendix

\section{Additional Implementation Details}

\subsection{Training}

We train all models using LLaMA-Factory \cite{zheng2024llamafactory} on 8 NVIDIA A100-SXM4-40G GPUs. We use full parameter fine-tuning across all the models in our experiment. Training hyperparameters are detailed in \Cref{tab:training_hyperparams}.
\begin{table}[ht]
\centering
\small
\begin{tabular}{ll}
\toprule
\textbf{Hyperparameter} & \textbf{Value} \\
\midrule
Precision & BF16 \\
Optimization & Flash Attention2 \\
Max Token Length & 8192 \\
Batch Size & 128 \\
Learning Rate & $5 \times 10^{-6}$ \\
LR Scheduler & Linear \\
Warmup Ratio & 0.03 \\
Weight Decay & 0.0 \\
Epochs & 1.0 \\
DeepSpeed & ZeRO-3 \\

\bottomrule
\end{tabular}
\caption{Training hyperparameters used for the experiments.}
\label{tab:training_hyperparams}
\end{table}

\subsection{Code Execution Filtering} 
\label{sec:code_exec_filter}
Before code execution, we filter out data instances where the solution code uses randomization libraries or the input is excessively large, to ensure deterministic and stable execution.

During code execution, to reduce execution time and computational overhead when executing codes at scale, we discard any data instance where code execution exceeds 5 seconds or results in a runtime error. Also, to avoid excessively long execution traces, we filter out execution traces with more than 300 lines.

\subsection{Data Generation Configuration}
\label{sec:data_gen_config}
We use the default generation configuration of the $\texttt{Qwen3-32B}$ translator. Specifically, we set the maximum token length to 16,382, with a temperature of 0.6, a top-p value of 0.95, and a top-k value of 20.

\subsection{Evaluation Configuration}
During evaluation, we set the temperature to 0.0 and use a maximum token length of 16,382. We enable the $\texttt{enable\_thinking=True}$ option to allow the model to think before generating solutions.


\subsection{Prompt Templates}
We present the prompt templates used in the experiment in \Cref{tab:prompt_template}. 

\section{Execution Trace Example}
\label{sec:execution_trace_example}
\Cref{tab:execution_trace_example} presents an example of executable code alongside its execution trace generated by $\texttt{Snoop}$. To enable tracing, the $\texttt{@snoop}$ decorator must be applied to the main entry function. The resulting execution trace includes function calls and return values, executed lines annotated with line numbers, and updated variable values. 

\section{Additional Experiment Results}
We present the experiment results for $\texttt{Qwen3-8B}$ in \Cref{tab:main_appendix}. Similar to the results of $\texttt{Qwen3-4B}$, our method overall outperforms all the baselines with particular strength in the coding benchmark.

Additionally, we present token length analysis of $\texttt{Qwen3-8B}$ on LiveBench in \Cref{tab:token_length_appendix}

\begin{table}[ht]
\centering
\scalebox{0.9}{
\begin{tabular}{lcc}
\toprule
\textbf{Method} & \textbf{Avg Token} & \textbf{Max Token Reached} \\
\midrule
No Training & 9030 & 116 \\
CodeI/O & 7362 & 83 \\
Ours    & 6289 & 54 \\
\bottomrule
\end{tabular}}
\caption{Token length statistics on the LiveBench evaluation using \texttt{Qwen3-8B}.}
\label{tab:token_length_appendix}
\end{table}

\section{Licenses}
We include the licenses of the datasets and models we used in this work.

Dataset License:
\begin{itemize}
  \item LiveBench: Apache-2.0 
\end{itemize}

Model Licenses:
\begin{itemize}
  \item $\texttt{Qwen3-4B}$:  Apache-2.0 
  \begin{itemize}
    \item \url{https://huggingface.co/Qwen/Qwen3-4B}
  \end{itemize}
  \item $\texttt{Qwen3-8B}$:  Apache-2.0 
  \begin{itemize}
    \item \url{https://huggingface.co/Qwen/Qwen3-8B}
  \end{itemize}
  \item $\texttt{Qwen3-32B}$: Apache-2.0 
  \begin{itemize}
      \item \url{https://huggingface.co/Qwen/Qwen3-32B}
  \end{itemize}
\end{itemize}

\begin{table*}[t]
\centering
\scalebox{0.9}{
\begin{tabular}{lcccccccc}
\toprule
\multirow{2}{*}{\textbf{Methods}} & \multicolumn{3}{c}{\textbf{LiveBench}} & \multirow{2}{*}{\textbf{MATH500}} & \multirow{2}{*}{\textbf{BBH}} & \multirow{2}{*}{\textbf{AGIEval}} & \multirow{2}{*}{\textbf{GPQA}} & \multirow{2}{*}{\textbf{Avg}} \\
\cmidrule(r){2-4}
& \textbf{Code} & \textbf{Math} & \textbf{Reasoning} & \\
\midrule
No Training & 45.8 & \textbf{64.4} & 67.1 & 88.0 & 74.4 & 31.5 & 36.3 & 58.2 \\
Code & 37.5 & 14.4 & 50.4 & 86.6 & 63.9 & \textbf{32.1} & 35.2 & 45.7 \\
Raw Trace & 33.5 & 20.6 & 22.8 & 86.0 & 64.5 & 31.2 & 33.7 & 41.7 \\
CodeI/O & 37.1 & 62.2 & \textbf{69.8} & 87.6 & 77.0 & 32.0 & 39.5 & 57.8 \\
Ours & \textbf{58.2} & 63.1 & 69.2 & \textbf{88.8} & \textbf{78.6} & 31.9 & \textbf{40.8} & \textbf{61.5} \\
\bottomrule
\end{tabular}}
\caption{Experiment results (accuracy) of \texttt{Qwen3-8B}. Bolded scores indicate the highest performance.}
\label{tab:main_appendix}
\end{table*}

 

\begin{table*}[ht]
\centering
\begin{tabular}{p{0.48\textwidth} | p{0.48\textwidth}}
\toprule
\textbf{Executable Code} & \textbf{Execution Trace generated by \texttt{Snoop}} \\
\midrule
\begin{lstlisting}[style=code]
import snoop # Import Snoop library

@snoop # Add the decorator to trace the function
def main_solution(num):
    if num < 0:
        return '-' + str(main_solution(-num))
    elif num < 7:
        return str(num)
    else:
        return str(main_solution(num // 7)) + str(num % 7)
main_solution(num=100) # Function call
\end{lstlisting}
&
\begin{lstlisting}[style=code]
>>> Call to main_solution # Function call 
 ...... num = 100 # Input 
   38 | def main_solution(num):
   39 |     if num < 0: # Executed line of code 
   41 |     elif num < 7:
   44 |         return str(main_solution(num // 7)) + str(num % 7)
     >>> Call to main_solution
     ...... num = 14 # Variable value
       38 | def main_solution(num):
       39 |     if num < 0:
       41 |     elif num < 7:
       44 |         return str(main_solution(num // 7)) + str(num % 7)
         >>> Call to main_solution 
         ...... num = 2
           38 | def main_solution(num):
           39 |     if num < 0:
           41 |     elif num < 7:
           42 |         return str(num)
         <<< Return value from main_solution: '2'
       44 |         return str(main_solution(num // 7)) + str(num % 7)
     <<< Return value from main_solution: '20'
   44 |         return str(main_solution(num // 7)) + str(num % 7)
 <<< Return value from main_solution: '202' # Return
\end{lstlisting}
\\
\bottomrule
\end{tabular}
\caption{An example of an execution trace generated by the Python tool called \texttt{Snoop}}
\label{tab:execution_trace_example}
\end{table*}

\begin{table*}[ht]
\small
\centering
\begin{tabular}
{m{0.95\linewidth}}
\toprule
\textbf{Execution Trace Translation Template} \\
\midrule
Given a question, an input to the question, and an execution trace that solves the question, your job is to translate the execution trace into a step-by-step thinking process. Here are some rules for translation:

- Use the exact values from the execution trace during the thought process to ensure the correctness of the thought process.

- Do not write code in your thinking process.

- Pretend you are not given the execution trace and you are solving the question by tracing the code by yourself. So, you should not mention that you are following the execution trace even when you are thinking. 

**Question** 

\{question\}

**Input**

\{input\}

**Execution Trace**

```

\{trace\}

```
\\
\midrule
\textbf{CodeI/O Solution Generation Template} \\
\midrule
You are given a question that requires some input and output variables as follows:

\{question\}

----

You are also given a solution code that solves the question:

\{code\}

----

Given the following input:

\{input\}

Predict the output of the question by tracing the given solution code step by step to reach the final output.
\\
\midrule
\textbf{User Prompt Template for Raw Execution Trace Baseline} \\
\midrule
You are given a question that requires some input and output variables as follows:

\{question\}

----

Here is the solution code that solves the question:

```
\{code\}
```

Given the following input:

\{input\}

Generate a step-by-step execution trace of by tracing the given solution code step by step to reach the final output.
\\
\midrule
\textbf{User Prompt Template for Code Generation Baseline} \\
\midrule
You are given a question that requires some input and output variables as follows:

\{question\}

----

Generate a solution code that solves the question. 
\\
\midrule
\textbf{User Prompt Template for CodeI/O and Our Method} \\
\midrule You are given a question that requires some input and output variables as follows:

\{question\}

----

You are also given a solution code that solves the question:

\{code\}

----

Given the following input:

\{input\}

Predict the output of the question by tracing the given solution code step by step to reach the final output.
\\
\bottomrule
\end{tabular}
\caption{Prompt Templates used in the experiments}
\label{tab:prompt_template}
\end{table*}

\lstset{
  style=plain,
  escapeinside={(*@}{@*)}
}
\begin{table*}[ht]
\centering
\begin{tabular}{p{0.48\textwidth} | p{0.48\textwidth}}
\toprule
\multicolumn{2}{p{0.92\textwidth}}{\textbf{Problem}: Given a string consisting of lowercase letters, what are all the possible unique permutations of the string?} \\
\midrule
\textbf{Our Solution} & \textbf{CodeI/O Solution} \\
\midrule
\begin{lstlisting}[style=plain]
To solve the problem of generating all unique permutations of the input string `'hrf'`, we follow these steps:

1. **Convert the string to a list of characters**:  
   The input string `'hrf'` is split into individual characters: `['h', 'r', 'f']`. This allows us to rearrange the characters systematically.

2. **Generate all permutations**:  
   Using a permutation algorithm or library function, we compute all possible orderings of the characters. For a 3-character string with unique elements, there are $3! = 6$ permutations. The permutations are:  
   - ('h', 'r', 'f')  
   - ('h', 'f', 'r')  
   - ('r', 'h', 'f')  
   - ('r', 'f', 'h')  
   - ('f', 'h', 'r')  
   - ('f', 'r', 'h')  

3. **Convert permutations to strings**:  
   Each permutation tuple is joined into a string:  
   - ('h', 'r', 'f') -> 'hrf'
   - ('h', 'f', 'r') -> 'hfr'  
   - ('r', 'h', 'f') -> 'rhf'  
   - ('r', 'f', 'h') -> 'rfh'  
   - ('f', 'h', 'r') -> 'fhr'
   - ('f', 'r', 'h') -> 'frh'  

4. **Return the final list**:  
   The resulting list of unique permutations is:  
   `['hrf', 'hfr', 'rhf', 'rfh', 'fhr', 'frh']`.  

This approach ensures all permutations are generated efficiently, leveraging the uniqueness of the input characters to avoid duplicates.
\end{lstlisting}
&
\begin{lstlisting}[style=plain]
The input string is 'hrf', which consists of three distinct lowercase letters. The solution code generates all possible permutations of these characters. Here's the step-by-step breakdown:

1. **Convert the input string to a list of characters**:  
   char_list = ['h', 'r', 'f']

2. **Generate permutations**:  
   Using itertools.permutations, the code calculates all possible orderings of the three unique characters. Since there are no duplicates, the total permutations are $3! = 6$:
   - ('h', 'r', 'f')
   - ('h', 'f', 'r')
   - ('r', 'h', 'f')
   - ('r', 'f', 'h')
   - ('f', 'h', 'r')
   - ('f', 'r', 'h')

3. **Convert tuples to strings**:  
   Each permutation tuple is joined into a string:
   - 'hrf'
   - 'hfr'
   (*@\textcolor{red}{- 'rhn'  >> ERROR: 'rhf'}@*)
   - 'rfh'
   - 'fhr'
   - 'frh'

4. **Final output**:  
   The list of unique permutations is returned as:
   
python
   ['hrf', 'hfr', 'rhf', 'rfh', 'fhr', 'frh']


**Answer:**  
The output is a list containing all 6 unique permutations of the input string 'hrf', as shown above.
\end{lstlisting}
\\
\bottomrule
\end{tabular}
\caption{An example of CodeI/O where the intermediate step is incorrect but the final output is correct. Our solution refers to the exact permutation values in the execution trace, which makes it easier to get the intermediate steps correct.}
\label{tab:case_study1}
\end{table*}

\lstset{
  style=plain,
  escapeinside={(*@}{@*)}
}
\begin{table*}[ht]
\centering
\begin{tabular}{p{0.48\textwidth} | p{0.48\textwidth}}
\toprule
\multicolumn{2}{p{0.92\textwidth}}{\textbf{Problem}: In a historical game of elimination, a group of people stand in a circle. Starting from a given person, every k-th person is eliminated until only one person remains. Given the number of people in the circle and the step count, which position in the circle will be the last remaining person? The input is \{'n': 17, 'k': 3\}} \\
\midrule
\textbf{Our Solution} & \textbf{CodeI/O Solution} \\
\midrule
\begin{lstlisting}[style=plain]
To solve the Josephus problem for `n = 17` and `k = 3`, we simulate the elimination process step by step. Here's the reasoning process:

1. **Initial Setup**:  
   The people are arranged in a circle with positions `[1, 2, 3, ..., 17]`. The elimination starts at position 1, and every 3rd person is removed.

2. **Recursive Elimination**:  
   The Josephus problem uses the formula:  
   $$
   J(n, k) = (J(n-1, k) + k) \mod n
   $$  
   where $ J(1, k) = 0 $ (0-based index). Since the problem uses 1-based indexing, the final result is $ J(n, k) + 1 $.

3. **Step-by-Step Calculation**:  
   Compute $ J(n, 3) $ for $ n = 1 $ to $ 17 $:  
   - $ J(1) = 0 $  
   - $ J(2) = (0 + 3) \mod 2 = 1 $  
   - $ J(3) = (1 + 3) \mod 3 = 1 $  
   - $ J(4) = (1 + 3) \mod 4 = 0 $  
   - $ J(5) = (0 + 3) \mod 5 = 3 $  
   - $ J(6) = (3 + 3) \mod 6 = 0 $  
   - $ J(7) = (0 + 3) \mod 7 = 3 $  
   - $ J(8) = (3 + 3) \mod 8 = 6 $  
   - $ J(9) = (6 + 3) \mod 9 = 0 $  
   - $ J(10) = (0 + 3) \mod 10 = 3 $  
   - $ J(11) = (3 + 3) \mod 11 = 6 $  
   - $ J(12) = (6 + 3) \mod 12 = 9 $  
   - $ J(13) = (9 + 3) \mod 13 = 12 $  
   - $ J(14) = (12 + 3) \mod 14 = 1 $  
   - $ J(15) = (1 + 3) \mod 15 = 4 $  
   - $ J(16) = (4 + 3) \mod 16 = 7 $  
   - $ J(17) = (7 + 3) \mod 17 = 10 $.  

   Adjusting to 1-based indexing: $ 10 + 1 = 11 $.

4. **Final Result**:  
   After all eliminations, the last remaining person is at position **11**.

**Answer**:  
The last remaining person is in position `11`.
\end{lstlisting}
&
\begin{lstlisting}[style=plain]
To solve the Josephus problem for `n = 17` and `k = 3`, we simulate the recursive elimination process as implemented in the provided code. The goal is to determine the position of the last remaining person in a circle of 17 individuals, where every 3rd person is eliminated in a clockwise fashion.

---

### **Understanding the Code Logic**

The recursive function `josephus_problem(array, k, index)` works as follows:

- **Base Case:** When only one person remains (`len(array) == 1`), return that person's position.
- **Recursive Step:**
    (*@\textcolor{red}{}@*)
  - Calculate the index of the next person to be eliminated using `(index (*@\textcolor{red}{+ k)}@*) % len(array)`.
  - Remove the person at that index.
  - Call the function recursively with the updated array and the new index (which is the index of the next person after the eliminated one).

In the `main_solution(n, k)` function:
- An array is initialized with positions from `1` to `n`.
- The recursive function is called with `k - 1` to adjust for zero-based indexing.

---

### **Step-by-Step Simulation**

We simulate the recursive calls manually for `n = 17`, `k = 3`, and track the eliminations:

1. Start with array: `[1, 2, 3, 4, 5, 6, 7, 8, 9, 10, 11, 12, 13, 14, 15, 16, 17]`, index = 0, k = 2.
2. Eliminate index `(0 + 2) % 17 = 2` -> remove `3`.
3. Eliminate index `(2 + 2) % 16 = 4` -> remove `6`.
...
16. Eliminate index `(3 + 2) % 3 = 0` -> remove `4`.
17. Eliminate index `(0 + 2) % 2 = 0` -> remove `11`.

At this point, only `13` remains.

---

### **Final Answer**
The last remaining person is at position (*@\textcolor{red}{**13**}@*).

\end{lstlisting}
\\
\bottomrule
\end{tabular}
\caption{An example of CodeI/O where an intermediate reasoning is wrong and the final answer is incorrect. In contrast, our solution correctly derives the formula, as shown in the execution trace, whereas CodeI/O uses an incorrect one.}
\label{tab:case_study2}
\end{table*}

\end{document}